\documentclass[a4paper]{llncs}
\usepackage{a4,a4wide}
\usepackage[latin1]{inputenc}
\usepackage{amssymb}
\usepackage{amsfonts}
\usepackage{multirow}
\usepackage{longtable}
\usepackage{booktabs}
\usepackage{enumerate}
\usepackage{amsmath}
\usepackage{bigdelim}
\usepackage{comment}

\newcommand{\f}[1]{\mathsf{#1}}
\newcommand{\n}[1]{\mbox{\textnormal{#1}}}
\newcommand{\subscrule}{\rule{0em}{1.45ex}}
\newcommand{\xforg}[2]{\f{forget}{}_{\subscrule #1}{#2}}
\newcommand{\xproj}[2]{\f{project}{}_{\subscrule #1}{#2}}

\newcommand{\xwsc}[2]{\f{gwsc}{}_{\subscrule #1}{#2}}
\newcommand{\true}{\top}
\newcommand{\false}{\bot}
\newcommand{\imp}{\rightarrow}
\newcommand{\revimp}{\leftarrow}

\newcommand{\subst}{\backslash}

\newcommand{\name}[1]{\emph{#1}}
\newcommand{\defname}[1]{\emph{#1}}

\newcommand{\entails}{\models}
\newcommand{\ALL}{\f{ALL}}

\newcommand{\q}[1]{\texttt{#1}}
\newcommand{\p}[1]{\textnormal{\textit{#1}}}

\newcommand{\introblock}[1]{

\medskip

\noindent
\hspace{0.7cm}
\begin{minipage}{13.3cm}
\it \indent #1
\end{minipage}

\medskip

}

\newcounter{codex}

\newenvironment{codex}
{\vskip 7pt plus1pt minus2pt
\refstepcounter{codex}%
\fontsize{9}{10}\selectfont%
\tt%
\noindent\hspace{1em}
\begin{tabular}[t]{l}}
{\end{tabular}\hspace{\fill}\textnormal{(\arabic{codex})}

\vskip 7pt plus1pt minus2pt
}

\renewcommand{\true}{\f{true}}
\renewcommand{\false}{\f{false}}

\newcommand{\parspecial}[1]{\paragraph{\textnormal{\textbf{#1}}}}

\title{Computing with Logic as Operator Elimination:\\ The
  ToyElim System\\[-0.55cm]}
\author{Christoph Wernhard}
\institute{Technische Universit\"{a}t Dresden\\
\small{\texttt{christoph.wernhard@tu-dresden.de}}}

\begin{document}
\maketitle

\begin{abstract}
A prototype system is described whose core functionality is, based on
propositional logic, the elimination of second-order operators, such as
Boolean quantifiers and operators for projection, forgetting and
circumscription.  This approach allows to express many representational and
computational tasks in knowledge representation -- for example computation of
abductive explanations and models with respect to logic programming semantics
-- in a uniform operational system, backed by a uniform classical semantic
framework.
\end{abstract}

\section{Computation with Logic as Operator Elimination}
\label{sec-intro}

We pursue an approach to computation with logic emerging from three theses:

\introblock{1. Classical first-order logic extended by some second-order
  operators suffices to express many techniques of knowledge representation.}

\noindent Like the standard logic operators, second-order operators can be
defined \name{semantically}, by specifying the requirements on an
interpretation to be a model of a formula whose principal functor is the
operator, depending only on semantic properties of the argument formulas.
Neither control structure imposed over formulas (e.g.\ Prolog), nor formula
transformations depending on a particular syntactic shape (e.g.\ Clark's
completion) are involved. Compared to classical first-order formulas, the
second-order operators give additional expressive power.  Circumscription is a
prominent knowledge representation technique that can be expressed with
second-order operators, in particular predicate quantifiers \cite{soqe}.

\introblock{2. Many computational tasks can be expressed as elimination of
  second-order operators.}

\noindent \name{Elimination} is a way to computationally process second-order
operators, for example Boolean quantifiers with respect to propositional
logic: The input is a formula which may contain the operator, for example a
quantified Boolean formula such as $\exists q\; ((p \revimp q) \land (q
\revimp r))$. The output is a formula that is equivalent to the input, but in
which the operator does not occur, such as, with respect to the formula above,
the propositional formula $p \revimp r$.
Let us assume that the method used to eliminate the Boolean quantifiers
returns formulas in which not just the quantifiers but also the quantified
propositional variables do not occur. This syntactic condition is usually met
by elimination procedures.  Our method then subsumes a variety of tasks:
Computation of uniform interpolants, QBF and SAT solving, as well as
computation of certain forms of abductive explanations, of propositional
circumscription, and of stable models, as will be outlined below.

\introblock{3. Depending on the application, outputs of computation with logic
  are conveniently represented by formulas meeting syntactic criteria.}

\noindent
If results of elimination are formulas characterized just up to semantics,
they may contain redundancies and be in a shape that is difficult to
comprehend. Thus, they should be subjected to simplification and canonization
procedures before passed to humans or machine clients.  The output format
depends on the application problem: What is a CNF of the formula?  Are certain
facts consequences of the formula? What are the models of the formula?  What
are its minimal models? What are its 3-valued models with respect to some
encoding into 2-valued logics?
Corresponding answers can be computed on the basis of normal form
representations of the elimination outputs: CNFs, DNFs, and full DNFs.
Of course, transformation into such normal forms might by itself be an
expensive task. Second-order operators allow to counter this by specifying a
small set of application relevant symbols that should be included in the
output (e.g.\ by Boolean quantification upon the irrelevant atoms).

\section{Features of the System}

\name{ToyElim}\footnote{http://cs.christophwernhard.com/provers/toyelim/,
  under GNU Public License.} is a prototype system developed to investigate
operator elimination from a pragmatic point of view with small
applications. For simplicity, it is based on propositional logic, although its
characteristic features should transfer to first-order logic.  It supports a
set of second-order operators that have been semantically defined in
\cite{cw-litproj,cw-circum,cw-logprog-short}.

\parspecial{Formula Syntax.}  As the system is implemented in Prolog, formulas
are represented by Prolog terms, the standard connectives corresponding to
\q{true}/0, \q{false}/0, \q{\~}/1, \q{,}/2, \q{;}/2, \q{->}/2, \q{<-}/2,
\q{<->}/2.  Propositional atoms are represented by Prolog atoms or compound
ground terms. The system supports propositional expansion with respect to
finite domains of formulas containing first-order quantifiers.

\parspecial{Forgetting.}  Existential Boolean quantification $\exists p\; F$
can be expressed as \defname{forgetting} \cite{cw-litproj,llm} in formula~$F$
about~atom $p$, written $\xforg{\{p\}}{(F)}$, represented by \q{forg([p],
  $F^\prime$)} in system syntax, where $F^\prime$ is the system representation
of~$F$.
To get an intuition of forgetting, consider the equivalence
$\xforg{\{p\}}{(F)} \equiv F[p\subst \true] \lor F[p\subst \false]$, where
$F[p\subst\true]$ ($F[p\subst\false]$) denotes $F$ with all occurrences of~$p$
replaced by $\true$ ($\false$). Rewriting with this equivalence constitutes a
naive method for eliminating the forgetting operator. The formula
$\xforg{\{p\}}{(F)}$ can be said to express the same as~$F$ about all other
atoms than $p$, but nothing about $p$.

\parspecial{Elimination and Pretty Printing of Formulas.}  The central
operation of the ToyElim system, elimination of second-order operators, is
performed by the predicate \q{elim(\p{F},\p{G})}, with input formula $F$ and
output formula $G$. For example, define as extension of \q{kb1}/1
a formula (after \cite{kakas:etal:93}) as follows:

\begin{codex}
\label{codex-kb-1}
kb1(((shoes\_are\_wet <- grass\_is\_wet),\\
\ \ \ \ \ (grass\_is\_wet <- rained\_last\_night),\\
\ \ \ \ \ (grass\_is\_wet <- sprinkler\_was\_on))).\\
\end{codex}

\noindent
After consulting this, we can execute the following query on the Prolog
toplevel:

\begin{codex}
\label{codex-q-forget}
?- kb1(F), elim(forg([grass\_is\_wet], F), G), ppr(G).\\
\end{codex}

\noindent
This results in binding \q{G} to the output of eliminating the forgetting
about \q{grass\_is\_wet}. The predicate~\q{ppr}/1 is one of several provided
predicates for converting formulas into application adequate shapes. It prints
its argument as CNF with clauses written as reverse implications:

\begin{codex}
\label{codex-elim-atomforg}
((shoes\_are\_wet <- rained\_last\_night),\\
\ (shoes\_are\_wet <- sprinkler\_was\_on))\n{.}
\end{codex}

\parspecial{Scopes.}  So far, the first argument of forgetting has been a
singleton set.  More generally, it can be an arbitrary set of atoms,
corresponding to nested existential quantification.  Even more generally, also
polarity can be considered: Forgetting can, for example, be applied only to
those occurrences of an atom which have negative polarity in a NNF formula.
This can be expressed by \name{literals} with explicitly written sign in the
first argument of the forgetting operator.  Forgetting about an atom is
equivalent to nested forgetting about the positive and the negative literal
with that atom.  In accord with this observation, we technically consider the
first argument of forgetting always as a \name{set of literals}, and regard an
unsigned atom there as a shorthand representing both of its literals.  For
example, $\q{[+grass\_is\_wet, shoes\_are\_wet]}$ is a shorthand for
$\q{[+grass\_is\_wet, +shoes\_are\_wet, -shoes\_are\_wet]}$.
Not just forgetting, but, as shown below, also other second-order operators
have a set of literals as parameter.  Hence, we refer to a set of literals in
this context by a special name, as \name{scope}.

\parspecial{Projection.}  In many applications it is useful to make explicit
not the scope that is ``forgotten'' about, but what is preserved.  The
\name{projection} \cite{cw-litproj} of formula~$F$ onto scope~$S$, which can
be defined for scopes~$S$ and formulas~$F$ as $\xproj{S}{(F)} \equiv
\xforg{\ALL - S}{(F)}$, where $\ALL$ denotes the set of all literals, serves
this purpose. Vice versa, forgetting could be defined in terms of projection:
$\xforg{S}{(F)} \equiv \xproj{\ALL - S}{(F)}.$ The call to \q{elim}/2 in the
query~(\ref{codex-q-forget}) can equivalently be expressed with projection
instead of forgetting by

\begin{codex}
elim(proj([shoes\_are\_wet, rained\_last\_night, sprinkler\_was\_on], F)\n{.}
\end{codex}

\parspecial{User Defined Logic Operators -- An Example of Abduction.}
ToyElim allows the user to specify macros for use in the input formulas of
\q{elim}/2. The following example extends the system by a logic operator
\q{gwsc} for a variant of the weakest necessary condition \cite{Lin:01:SNC},
characterized in terms of projection:

\begin{codex}
:- define\_elim\_macro(gwsc(S, F, G), \~{ }proj(complements(S), (F, \~{ }G))).
\end{codex}

\noindent
Here \q{complements(S)} specifies the set of the literal complements of the
members of the scope specified by \q{S}.  The term \q{gwsc(S, F, G)} is the
system syntax for $\xwsc{S}{(F,G)}$, the \name{globally weakest sufficient
  condition} of formula~$G$ on scope~$S$ within formula~$F$, which satisfies
the following: A formula $H$ is equivalent to $\xwsc{S}{(F,G)}$ if and only if
it holds that (1.) $H \equiv \xproj{S}{(H)}$; (2.) $F \entails H \imp
G$; (3.)  For all formulas $H^\prime$ such that $H^\prime \equiv
\xproj{S}{(H^\prime)}$ and $F \entails G \imp H^\prime$ it holds that $H
\entails H^\prime$.  
%
%
With the \q{gwsc} operator certain abductive tasks \cite{kakas:etal:93} can
be expressed.  The following query, for example, yields abductive explanations
for \q{shoes\_are\_wet} 
in terms of $\{$\q{rained\_last\_night}, \q{sprinkler\_was\_on}$\}$
with respect to the knowledge base~(\ref{codex-kb-1}):

\begin{codex}
\label{codex-intro-ppm}
?- kb1(F),\\
\ \ \ elim(gwsc([rained\_last\_night, sprinkler\_was\_on],
		  F,		  
		  shoes\_are\_wet),\\
\ \ \ \ \ \ \ \ G),\\
\ \ \	ppm(G).
\end{codex}

\noindent
The predicate \q{ppm}/1 serves, like \q{ppr}/1, to convert formulas to
application adequate shape.  It writes a DNF of its input, in list notation,
and simplified such that it does not contain tautologies and subsumed clauses.
In the example the output has two clauses, each representing an alternate
explanation:

\begin{codex}
[[rained\_last\_night],[sprinkler\_was\_on]]\n{.}
\end{codex}

\parspecial{Scope-Determined Circumscription.}  A further second-order
operator supported by ToyElim is \name{scope-determined circumscription}
\cite{cw-circum}.  The corresponding functor \q{circ} has, like \q{proj} and
\q{forg}, a scope specifier and a formula as arguments.  It allows to express
\name{parallel predicate circumscription with varied predicates}
\cite{circumlif} (only propositional, since the system is based on
propositional logic).  The scope specifier controls the effect of
circumscription: Atoms that occur just in a \emph{positive} literal in the
scope are minimized; symmetrically, atoms that occur just \emph{negatively}
are maximized; atoms that occur in \emph{both polarities} are fixed; and atoms
that do \emph{not at all} occur in the scope are allowed to vary.  For
example, the scope specifier, $\q{[+abnormal, bird]}$, a shorthand for
$\q{[+abnormal, +bird, -bird]}$, expresses that $\q{abnormal}$ is minimized,
$\q{bird}$ is fixed, and all other predicates are varied.

\parspecial{Predicate Groups and Systematic Renaming.}  Semantics for
knowledge representation sometimes involve what might be described as handling
different occurrences of a predicate differently -- for example depending on
whether it is subject to negation as failure.
If such semantics are to be modeled with classical logic, then these
occurrences can be identified by using distinguished predicates, which are
equated with the original ones when required.
To this end, ToyElim supports the handling of \name{predicate groups}: The
idea is that each predicate actually is represented by several
\name{corresponding} predicates~$p^0, \ldots, p^n$, where the superscripted
index is called \name{predicate group}.  In the system syntax, the predicate
group of an atom is represented within its main functor: If the group is
larger than~$0$, the main functor is suffixed by the group number; if it is
$0$, the main functor does not end in a number. For example $p(a)^0$ and
$p(a)^1$ are represented by \q{p(a)} and \q{p1(a)}, respectively.
In scope specifiers, a number is used as shorthand for the set of all literals
whose atom is from the indicated group, and a number in a sign functor for the
set of those literals which have that sign and whose atom is from the
indicated group.  For example, \q{[+(0), 1]} denotes the union of the set of
all positive literals whose atom is from group 0 and of the set of all
literals whose atom is from group 1.
Systematic renaming of all atoms in a formula that have a specific group to
their correspondents from another group can be expressed in terms of
forgetting \cite{cw-logprog-short}. The \mbox{ToyElim} system provides the
second-order operator~\q{rename} for this. For example, \q{rename([1-0], $F$)}
is equivalent to $F$ after eliminating second-order operators, followed by
replacing all atoms from group~1 with their correspondents from group~0.

\parspecial{An Example of Modeling a Logic Programming Semantics.}
Scope-determined circumscription and predicate groups can be used to express
the characterization of the stable models semantics in terms of
circumscription \cite{Lin:Thesis} (described also in
\cite{Lifschitz:08:12defs,cw-logprog-short}).  Consider the following
knowledge base:

\begin{codex}
\label{codex-kb-2}
kb2(((shoes\_are\_wet <- grass\_is\_wet),\\
\ \ \ \ \ (grass\_is\_wet <- sprinkler\_was\_on, \~{ }sprinkler\_was\_abnormal1),\\
\ \ \ \ \ sprinkler\_was\_on)).
\end{codex}

\noindent
Group~1 is used here to indicate atoms that are subject to negation as
failure: All atoms in (\ref{codex-kb-2}) are from group~0, except for
\q{sprinkler\_was\_abnormal1}, which is from~1.  The user defined operator
\q{stable} renders the stable models semantics:

\begin{codex}
:- define\_elim\_macro(stable(F), rename([1-0], circ([+(0),1], F))).
\end{codex}

\noindent
The following query then yields the stable models:

\begin{codex}
:- kb2(F), elim(stable((F)), G), ppm(G).
\end{codex}

\noindent
The result is displayed with \q{ppm}/1, as in query~(\ref{codex-intro-ppm}).
It shows here a DNF with a single clause, representing a single model. The
positive members of the clause constitute the answer set

\begin{codex}
[[grass\_is\_wet, shoes\_are\_wet, \~{ }sprinkler\_was\_abnormal,
 sprinkler\_was\_on]]\n{.}
\end{codex}

\noindent
If it is only of interest whether \q{shoes\_are\_wet} is a consequence of the
knowledge base under stable models semantics, projection can be applied to
obtain a smaller result. The query

\begin{codex}
:- kb2(F), elim(proj([shoes\_are\_wet], stable(F)), G),	ppm(G).
\end{codex}

\noindent
will effect that the DNF \q{[[shoes\_are\_wet]]} is printed.

\section{Implementation}
\label{sec-meth}

The ToyElim system is implemented in SWI-Prolog and can invoke external
systems such as SAT and QBF solvers.  It runs embedded in the Prolog
environment, allowing for example to pass intermediate results between its
components through Prolog variables, as exemplified by the queries shown
above.

The implementation of the core predicate \q{elim}/2 maintains a formula which
is gradually rewritten until it contains no more second-order operators.  It
is initialized with the input formula, preprocessed such that only two
primitively supported second-order operators remain: forgetting and renaming.
It then proceeds in a loop where alternately equivalence preserving
simplifying rewritings are applied, and a subformula is picked and handed over
for elimination to a specialized procedure.  The simplifying rewritings
include distribution of forgetting over subformulas and elimination steps that
can be performed with low cost \cite{cw-tableaux}.  Rewriting of subformulas
with the Shannon expansion enables low-cost elimination steps. It is
performed at this stage if the expansion, combined with low-cost elimination
steps and simplifications, does not lead to an increase of the formula size.
The subformula for handing over to a specialized method is picked with the
following priority: First, an application of forgetting upon the whole
signature of a propositional argument, which can be reduced by a SAT solver to
either $\true$ or $\false$, is searched. Second, a subformula that can be
reduced analogously by a QBF solver, and finally a subformula which properly
requires elimination of forgetting. For the latter, ToyElim schedules a
portfolio of different methods, where currently two algorithmic approaches are
supported: Resolvent generation (SCAN, Davis-Putnam method) and rewriting of
subformulas with the Shannon expansion
\cite{Murray:03:DNNF,cw-tableaux}. Recent SAT preprocessors partially perform
variable elimination by resolvent generation.  \name{Coprocessor}
\cite{coprocessor} is such a preprocessor that is configurable such that it
can be invoked by ToyElim for the purpose of performing the elimination of
forgetting.

\section{Conclusion}
\label{sec-conc}

We have seen a prototype system for computation with logic as elimination of
second-order operators.  The system helped to concretize requirements on the
user interface and on processing methods of systems which are entailed by that
approach.  In the long run, such a system should be based on more expressive
logics than propositional logic.  ToyElim is just a first pragmatic attempt,
taking advantage of recent advances in SAT solving.  A major difference in a
first-order setting is that computations of elimination tasks then inherently
do not terminate for all inputs.

A general system should for special subtasks not behave worse than systems
specialized for these.  This can be achieved by identifying such subtasks, or
by general methods that implicitly operate like the specialized ones. ToyElim
identifies SAT and QBF subtasks.  It is a challenge to extend this range, for
example, such that the encoded stable model computation would be performed
efficiently.
The system picks in each round a single subtask that is passed to a
specialized solver.  We plan to experiment with a more flexible regime, where
different subtasks are alternately tried with increasing timeouts.

Research on the improvement of elimination methods includes further
consideration of techniques from SAT preprocessors, investigation of tableau
and DPLL-like techniques \cite{cw-tableaux,Darwiche:05:sat-kc}, and, in the
context of first-order logic, the so called \name{direct methods} \cite{soqe}.
In addition, it seems worth to investigate further types of output:
incremental construction, like enumeration of model representations, and
representations of proofs.

The approach of computation with logic by elimination leads to a system that
provides a uniform user interface covering many tasks, like satisfiability
checking, computation of abductive explanations and computation of models for
various logic programming semantics.  Variants of established concepts can be
easily expressed on a clean semantic basis and made operational.  The approach
supports the co-existence of different knowledge representation techniques in
a single system, backed by a single classical semantic framework.  This seems
a necessary precondition for logic libraries that accumulate knowledge
independently of some particular application.

\bibliographystyle{abbrv}
\bibliography{bibclelim}

\end{document}